\documentclass{article} 
\usepackage{iclr2026_conference,times}

\usepackage{amsmath,amsfonts,bm}

\def\eqref#1{equation~\ref{#1}}

\def\1{\bm{1}}

\DeclareMathAlphabet{\mathsfit}{\encodingdefault}{\sfdefault}{m}{sl}
\SetMathAlphabet{\mathsfit}{bold}{\encodingdefault}{\sfdefault}{bx}{n}

\usepackage{hyperref}
\usepackage{url}

\usepackage[utf8]{inputenc}

\usepackage{makecell}      
\usepackage[table]{xcolor} 
\usepackage[utf8]{inputenc}
\usepackage{amsmath}           
\usepackage{graphicx}          
\usepackage{booktabs}      
\usepackage{enumitem}
\usepackage{subcaption} 

\definecolor{lightgray}{gray}{0.9}

\newcommand{\equal}{\textsuperscript{*}}           
\newcommand{\leader}{\textsuperscript{$\dagger$}}  
\newcommand{\corr}{\textsuperscript{$\ddagger$}}   
\newcommand{\intern}{\textsuperscript{$\S$}}       

\title{FantasyWorld: Geometry-Consistent World Modeling via Unified Video and 3D Prediction}

\author{
    Yixiang Dai\textsuperscript{1}\equal\intern,\quad
    Fan Jiang\textsuperscript{1}\equal\leader\corr,\quad
    Chiyu Wang\textsuperscript{1}\equal,\quad
    Mu Xu\textsuperscript{1},\quad
    Yonggang Qi\textsuperscript{2}\corr \\
    \textsuperscript{1}AMAP, Alibaba Group,\quad
    \textsuperscript{2}Beijing University of Posts and Telecommunications \\
    {\footnotesize\texttt{\{daiyixiang.dyx,~frank.jf,~wangchiyu.wcy,~xumu.xm\}@alibaba-inc.com}}\\
    {\footnotesize\texttt{qiyg@bupt.edu.cn}}
}

\iclrfinalcopy 
\begin{document}

\maketitle

\begingroup
\renewcommand{\thefootnote}{\fnsymbol{footnote}}
\footnotetext[1]{Equal contribution.}
\footnotetext[2]{Project leader.}
\footnotetext[3]{Corresponding author.}
\footnotetext[4]{Work done during internship at AMAP, Alibaba Group.}
\endgroup

\begin{abstract}

High-quality 3D world models are pivotal for embodied intelligence and Artificial General Intelligence (AGI), underpinning applications such as AR/VR content creation and robotic navigation. 
Despite the established strong imaginative priors, current video foundation models lack explicit 3D grounding capabilities, thus being limited in both spatial consistency and their utility for downstream 3D reasoning tasks. 
In this work, we present \textsc{FantasyWorld}, a geometry-enhanced framework that augments frozen video foundation models with a trainable geometric branch, enabling joint modeling of video latents and an implicit 3D field in a single forward pass. 
Our approach introduces cross-branch supervision, where geometry cues guide video generation and video priors regularize 3D prediction, thus yielding consistent and generalizable 3D-aware video representations. 
Notably, the resulting latents from the geometric branch can potentially serve as versatile representations for downstream 3D tasks such as novel view synthesis and navigation, without requiring per-scene optimization or fine-tuning. 
Extensive experiments show that \textsc{FantasyWorld} effectively bridges video imagination and 3D perception, outperforming recent geometry-consistent baselines in multi-view coherence and style consistency. 
Ablation studies further confirm that these gains stem from the unified backbone and cross-branch information exchange.

\end{abstract}

\begin{figure}[h]
  \centering
  \includegraphics[width=1\linewidth]{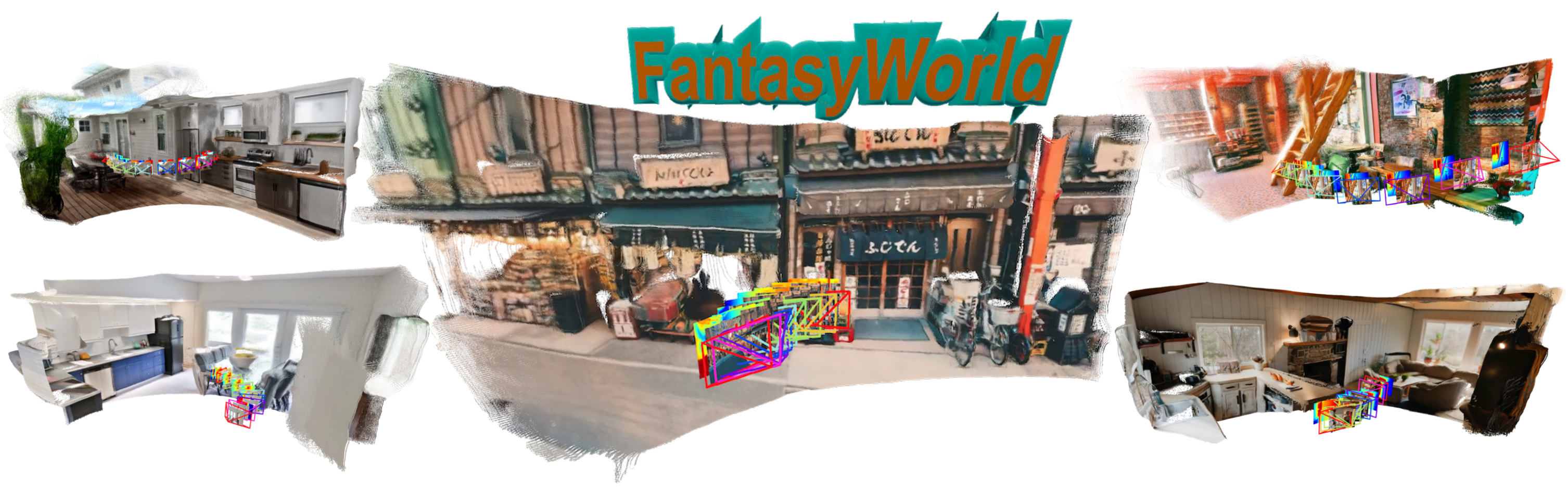}
  \caption{\textsc{FantasyWorld} overview. 
Given multimodal inputs (image, text, and camera trajectory), the model generates photorealistic videos along the specified views while constructing an implicit 3D representation for consistent geometry.}
  \label{fig:overview}
\end{figure}

\section{Introduction}

Building world models has long been viewed as a key step toward Artificial General Intelligence (AGI). 
By modeling environments, objects, causal relations, and temporal dynamics, world models enable agents to predict, plan, and generate, endowing them with human-like understanding and creative abilities. 
At the core of this vision lies the construction of high-quality, diverse 3D environments, which support a wide spectrum of applications spanning AR/VR content creation, robotic navigation, and embodied AI at large.
Plenty of approaches have been explored towards the goal of generating coherent 3D scenes, involving leveraging 2D generative priors to enable 3D generation \citep{chen2023fantasia3d,bahmani20244d} via SDS \citep{poole2022dreamfusion}, building entire 3D scenes by multi-view images with NeRF \citep{mildenhall2021nerf} and 3DGS \citep{kerbl20233d}. 

Recently, camera-guided video diffusion models have gained popularity, where image frames implicitly encode the scene’s 3D structure with multiple images from different viewpoints, as videos are natural 2D projections of dynamic 3D worlds. 
Such methods harness the spatiotemporal capacity of video diffusion to improve 3D consistency and multi-view modeling, such as ReconX \citep{liu2024reconx}, Gen3C \citep{ren2025gen3c}, DimensionX \citep{sun2024dimensionx}, and ViewCrafter \citep{yu2024viewcrafter}, to name a few. 
These models excel at producing geometrically plausible local shapes and mitigating view inconsistency through temporal priors. 

However, despite being trained on vast Internet video corpora and exhibiting strong imaginative priors, generative video models lack explicit 3D supervision, making it difficult to preserve real-world structure and spatial consistency. 
This reveals a central challenge in world model construction: \emph{how to inject reliable geometric grounding into video generation without sacrificing creative capacity}. 

Lately, the emergence of 3D foundation models, such as DUSt3R \citep{wang2024dust3r}, MASt3R \citep{duisterhof2025mast3r}, Fast3R \citep{yang2025fast3r}, and VGGT \citep{wang2025vggt}, demonstrates that robust geometry can be predicted in a single forward pass without additional 3D reconstruction, providing a scalable solution to geometry-consistent generation. 
Consequently, several works attempt to couple 2D video generation with 3D reasoning. 
Voyager \citep{huang_voyager_2025} trains an end-to-end model that jointly predicts RGB and depth while maintaining world consistency through a cache and geometry-injected frames, and then reconstructs explorable 3D scenes.
Geometry Forcing \citep{wu_geometry_2025} further marries video diffusion with an explicit 3D representation to strengthen geometric consistency during training.
Matrix-3D \citep{yang_matrix-3d_2025} shows two practical routes from omni-directional video to a navigable world, including an optimization path and a large panorama reconstruction model that directly infers 3D Gaussians from video latents. 
WonderWorld \citep{yu_wonderworld_2025} focuses on interactive single-image scene authoring with layered Gaussian surfels and guided depth diffusion for fast, connected scene creation.
Vidu4D \citep{wang_vidu4d_2024} shows that even a single generated video can support high-quality 4D reconstruction when paired with dynamic Gaussian surface elements. 
Complementary to these trends in content creation, GaussianWorld \citep{zuo_gaussianworld_2025} frames perception as streaming 4D occupancy forecasting, underscoring the value of a 3D world representation that can generalize across embodied tasks. 

Despite significant progress in video-to-3D modeling, several limitations remain. 
First, most video generative models operate purely within the video domain, yielding features that cannot directly support 3D reasoning. 
When explicit 3D reconstruction is desired, they often resort to additional scene-specific optimization with NeRF \citep{mildenhall2021nerf} or 3DGS \citep{kerbl20233d}, which introduces computation overhead. 
Second, video imagination and 3D perception remain weakly coupled at inference, preventing mutual reinforcement, which effectiveness is evidenced by recent works in 3D scene understanding \citep{huang2025mllms}. 
For instance, although Voyager \citep{huang_voyager_2025} predicts RGB and depth jointly to maintain world consistency, and WorldExplorer \citep{schneider2025worldexplorer} leverages video-based imagination for scene exploration, in both cases the two processes operate largely independently, highlighting the persistent limitation of weak coupling between video generation and 3D perception.
Third, many approaches, such as Geometry Forcing \citep{wu_geometry_2025}, integrate 3D priors by fine-tuning video foundation models (VFMs) while keeping a 3D model like VGGT frozen, which incurs substantial computational cost and risks compromising the VFM’s general generative capacity. 

To address these limitations, we introduce \textsc{FantasyWorld}, a geometry-enhanced framework that efficiently produces reusable 3D-consistent features by augmenting frozen VFMs with an additional trainable branch for geometric inference, tightly coupling video imagination and 3D perception without expensive per-scene optimization or fine-tuning. 
Instead of predicting depth or point clouds from RGB images, we directly infer camera parameters and 3D signals from video latents. 
This is inspired by VGGT but achieves tighter integration between generative and geometric modeling. 
Concretely, we split the backbone of video foundation models (i.e., Wan2.1 in our case) into Preconditioning Blocks (PCB) that inject video priors and stabilize latents, and Integrated Reconstruction and Generation Blocks (IRG) that fuse spatiotemporal tokens with a geometry co-encoder to predict a geometry-aware implicit 3D field. 
As a result, our model generates camera-conditioned video features alongside an explicit 3D representation in a single forward pass, without relying on additional 3D reconstruction (e.g., NeRF or 3DGS) or iterative memory refinement as in Voyager (\cite{huang_voyager_2025}).

To this end, our contributions are as follows: (i) Unified video-3D modeling: We propose \textsc{FantasyWorld}, a geometry-enhanced framework that jointly predicts video latents and an implicit 3D field through a single backbone, preserving imaginative priors while exposing explicit geometry. 
(ii) 2D/3D cross-branch supervision: We introduce constraints that let geometry supervise video features and video priors regularize 3D prediction, ensuring 3D-consistent frames inference. 
(iii) Potential for generalizable 3D features: We expect that the resulting video-3D representations serve as versatile features for downstream tasks, such as novel view synthesis and navigation, without per-task adaptation, which has been evidenced by recent works, such as AnySplat \citep{jiang_anysplat_2025}.

\section{Related Work}
\subsection{Feed-Forward Reconstruction}
Feed-forward reconstruction methods have achieved promising results in recovering 3D properties of a scene from a set of images in a single pass \citep{wang_dust3r_2024, wang2025continuous, wang20243d, duisterhof2025mast3r, tang2025mv, yang_fast3r_2025, zhang2024monst3r, zhang2025flare, wang2025vggt, wang2025pi}. 
DUSt3R~\citep{wang_dust3r_2024} and MASt3R~\citep{duisterhof2025mast3r} directly estimate point clouds from images, which is suitable for challenging scenarios such as low-texture regions, but can only input two images at a time.
Fast3R~\citep{yang_fast3r_2025} can process thousands of frames at once, surpassing the aforementioned restrictions.
VGGT~\citep{wang2025vggt} introduces a more generalized framework capable of supporting input from one frame to multiple frames and producingg multiple 3D attributes. 
Furthermore, its feature backbone serves as a versatile feature extractor for various downstream tasks. Recently, many methods have integrated diffusion knowledge into reconstruction techniques, thereby enriching the 3D modeling process with the ability to imagine occluded or unobserved areas~\citep{fu2024geowizard, hu2025depthcrafter, ke2024repurposing, lu2025matrix3d, yang2024unipad, zhu2024spa, zhu2023ponderv2, liang2025wonderland}.
In this work, we adopt an architecture similar to VGGT as our 3D implicit feature extractor. 
Specifically, we extract implicit 3D features from WanDiT block and integrate them within a multi-task learning paradigm.

\subsection{Geometry-aware Video Generation}
The consistency of generated videos with the real physical world is a crucial challenge in simulation. 
Many previous works have attempted to incorporate geometric constraints during the video generation process, primarily categorized into explicit guidance and implicit guidance. 
For explicit guidance, many prior works explicitly incorporate 3D signals (such as point clouds, mesh, etc.) obtained from the first frame into the model, thereby demonstrably improving generation consistency~\citep{yu2024wonderjourney, huang_voyager_2025, yang2025matrix, cao2025uni3c}. 
These methods often suffer from insufficient point cloud accuracy and limited scope when subjected to significant viewpoint changes. 
For implicit guidance, many prior works investigate methods to enable models to perceive 3D structural information within the diffusion process~\citep{zhang2025world, team_aether_2025, wu_geometry_2025}. 
Geometry-Forcing~\citep{wu_geometry_2025} aligns the model's intermediate representations with the features of a pre-trained geometric foundational model, risks undermining the creativity of large-scale video generation models trained on massive video data.

\subsection{Joint 3D and Video Generation}
Integrating both video and 3D structural information enables the generation of comprehensive 3D scenes for applications in embodied AI and simultaneously facilitates a better representation of the underlying physical world. 
AETHER~\citep{team_aether_2025} unifies RGB–depth modeling and couples reconstruction with video generation.
DeepVerse~\citep{chen2025deepverse} achieves interactive joint generation via an autoregressive paradigm and control via text-specified control signals. 
Voyager~\citep{huang_voyager_2025} incorporates point cloud projection within its joint generation framework to achieve camera control. 

In this paper, we introduce an implicit 3D feature encoding branch designed to decode various 3D attributes. 
Through mutual enhancement achieved during the diffusion process, it simultaneously yields video features and 3D structural features.

\section{Methodology}
\label{sec:method}

\subsection{Overview}
\label{sec:overview}
\textsc{FantasyWorld} is a unified feed-forward model for joint video and 3D scene generation. 
Given a reference image, an optional text prompt, and a target camera trajectory, the model produces a video aligned with the specified views while simultaneously constructing an implicit 3D representation. 
Inputs are encoded by pretrained backbones: CLIP~\citep{radford2021learning} for images, umT5~\citep{chung2023unimax} for text, and a learned camera encoder, following Wan’s Plücker-ray design~\citep{wan2025wan}, for camera poses. 
These signals jointly condition both the video and geometry branches during training and inference.

\begin{figure}[t]
  \centering
  \includegraphics[width=1\linewidth]{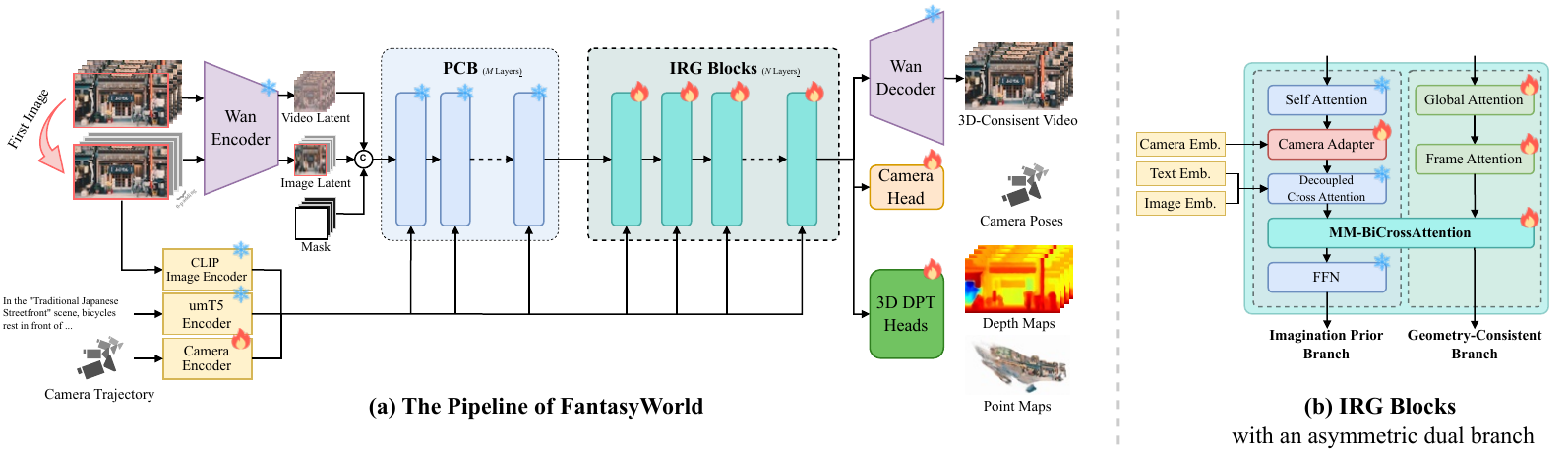}
  \caption{Overview of \textsc{FantasyWorld}. Inputs (image, text, camera) are processed by PCBs and stacked IRG blocks, where an asymmetric dual-branch design couples video synthesis with 3D reasoning. The model outputs geometry-consistent video frames and task-agnostic 3D features.}
  \label{fig:network}
\end{figure}

As shown in Fig.~\ref{fig:network}~(a), the front end employs \emph{Preconditioning Blocks (PCBs)} that reuse the frozen WanDiT denoiser to supply partially denoised latents, ensuring the geometry pathway operates on meaningful features rather than pure noise. 
The backbone then consists of stacked \emph{Integrated Reconstruction and Generation (IRG) Blocks}, which iteratively refine video latents and geometry features under multimodal conditioning. 
Each IRG block contains an asymmetric dual-branch structure (Fig.~\ref{fig:network}~(b)): an \emph{Imagination Prior Branch} for appearance synthesis and a \emph{Geometry-Consistent Branch} for explicit 3D reasoning, coupled through lightweight adapters and cross attention.

The outputs are twofold. 
The imagination branch generates geometry-consistent video frames along the trajectory, while the geometry branch produces task-agnostic 3D features decoded by DPT heads into depth maps, point maps, and camera poses. 
This unified design supports downstream tasks such as novel view synthesis, pose estimation, and depth prediction, all without per-scene optimization.

In summary, \textsc{FantasyWorld} bridges generative video priors with structured geometric reasoning in a single forward pass, producing outputs that are both photorealistic and geometrically consistent.

\subsection{Preconditioning Blocks}
\label{sec:pcb}

\begin{figure}[h]
  \centering
  \includegraphics[width=1\linewidth]{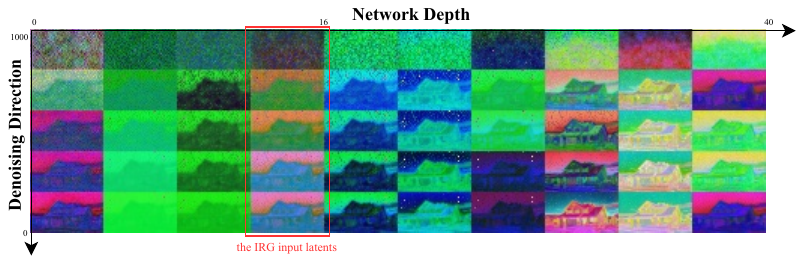}
  \caption{PCA over timestep–block pairs: rows vary timesteps top $\rightarrow$ bottom, columns vary blocks left $\rightarrow$ right; the red rectangle marks the IRG input latents.}
  \label{fig:method:denoised}
\end{figure}

Denoising diffusion models progressively remove noise across timesteps, revealing structure and details in the signal. 
Recent theory~\citep{han2025feature} shows that the denoising objective balances learning signal and noise, enabling structural information to emerge gradually—consistent with the empirical observation that features become more informative as denoising unfolds. 
We further observe a similar effect along network depth: even at a fixed timestep, deeper WanDiT layers produce clearer spatial structure (Fig.~\ref{fig:method:denoised}), suggesting that denoising progresses across both time and depth.

Motivated by this, we introduce \emph{Preconditioning Blocks (PCBs)} at the front end of our framework. 
We reuse the first 16 frozen layers of Wan2.1, following its camera conditioning design, to partially denoise video latents. 
This ensures that inputs to the geometry branch contain geometry-relevant cues rather than pure noise, reducing gradient variance and avoiding early training dominated by high-noise latents. 
PCBs thus bridge noisy initialization and geometry-aware processing, providing stable supervision from the start and allowing the geometry branch to focus on refining structure.

\subsection{Integrated Reconstruction and Generation Blocks}

At the core of \textsc{FantasyWorld} are the \emph{Integrated Reconstruction and Generation (IRG) Blocks}, stacked as the fundamental units of the model. 
Each IRG block adopts an asymmetric dual-branch design (Fig.~\ref{fig:network}~(b)): the \emph{Imagination Prior Branch} reuses the pretrained Wan2.1 backbone to propagate appearance-rich spatiotemporal features, while the \emph{Geometry-Consistent Branch} projects them into a geometry-aligned latent space. 
Unlike VGGT, which relied on DINO features, we bridge the geometry pathway to Wan latents, ensuring geometry is inferred in the same domain as video synthesis and avoiding feature mismatch.

IRG blocks unify the two branches through bidirectional cross-attention (MM-BiCrossAttn~\citep{liu2024grounding,liu2025javisdit}). 
Here, geometric cues regularize video features for multi-view coherence, while video priors provide imaginative signals that complete occluded regions and refine geometry. 
As cooperative units, stacked IRG blocks progressively enhance both video latents and geometry features, forming the core mechanism where imagination and structure converge.

The geometry branch outputs an implicit representation decoded by a custom 3D DPT head. 
This head performs temporal decoding aligned with the WanVAE video frames. 
Motivated by the depth-wise emergence of structure observed in Sec.~\ref{sec:pcb}, the head inverts the conventional reassemble strategy: instead of drawing fine features from early encoder layers dominated by noise, it sources them from late diffusion blocks where semantics are stronger and denoising is mature. 
Anchoring predictions in these stable features improves depth accuracy, stabilizes pose estimation, and enhances the consistency of the implicit 3D field.

\subsection{Bridge to Unify Training}
\label{sec:training}
\textbf{Two-stage framework.} 
Training in \textsc{FantasyWorld} follows two stages: first \emph{bridging} the geometry branch to the Wan feature space, then \emph{unifying} geometry and video through bidirectional cross-attention. 
Stage 1 freezes the Wan2.1 backbone and adapts the geometry branch to consume hidden features; Stage 2 introduces cross-branch adapters for joint optimization, while keeping the backbone frozen. Please refer to~\ref{app:camera_control},~\ref{app:dpt3d}~\ref{app:training} for more implementation and training details.

\textbf{Stage 1: Latent Bridging.} 
We select hidden features from block 16 of Wan2.1 and feed them to the geometry branch through a lightweight transformer adapter that maps to a geometry-aligned latent space. 
This latent is encoded and decoded into camera, depth, and point map predictions, while only the geometry branch is trained. 
Supervision is applied via:
\[
\mathcal{L}_{\mathrm{geo}} = \alpha\,\mathcal{L}_{\mathrm{depth}} + \beta\,\mathcal{L}_{\mathrm{pmap}} + \gamma\,\mathcal{L}_{\mathrm{camera}}
\]
where $\mathcal{L}_{\mathrm{depth}}$ follows Video Depth Anything~\citep{chen2025video}, $\mathcal{L}_{\mathrm{pmap}}$ follows VGGT~\citep{wang2025vggt}, and $\mathcal{L}_{\mathrm{camera}}$ is a Huber penalty. 
This stage ensures the geometry branch operates stably on Wan latents instead of raw noise.

\textbf{Stage 2: Unified Co-Optimization.} 
From block 16 onward, we insert one bidirectional cross-attention adapter after each of 24 transformer blocks, aligned with the geometry branch. 
For video tokens $X_v$ and geometry tokens $X_g$, with projections $(Q_v,K_v,V_v)$ and $(Q_g,K_g,V_g)$, attention is:
\[
A = \mathrm{softmax}\!\left(\tfrac{Q_v K_g^{\top}}{\sqrt{d_k}}\right)
\]
and updates are:
\[
X_v^{+} = X_v + \gamma_v A V_g, \qquad
X_g^{+} = X_g + \gamma_g A^{\top} V_v
\]
with learnable gates $\gamma_v,\gamma_g$. 
Geometry-to-video updates enforce 3D consistency, while video-to-geometry updates inject generative priors. 
Camera parameters are embedded with a pose encoder~\citep{wan2025wan}; we modify the pose adapter to predict only the shift $\beta_i$, injected additively:
\[
f_i = f_{i-1} + \beta_i,
\]
applied to the first 24 of 40 blocks.

\textbf{Training objective.} 
The final objective is:
\[
\mathcal{L}_{\mathrm{total}} = 
\mathbb{E}_{z_0,\epsilon,t,c} 
\Big[ \|\epsilon_{\theta}(z_t,t,c) - \epsilon\|_2^2 \Big] 
+ \lambda \mathcal{L}_{\mathrm{geo}}
\]
combining the standard diffusion loss with geometry supervision which aggregates depth, point map, and camera supervision. 
The weight $\lambda$ balances video generation and geometry learning, enforcing multi-view coherence and enabling cross-branch co-adaptation.

\section{Experiment} 
We conduct comprehensive experiments to evaluate \textsc{FantasyWorld} across large-scale benchmarks and diverse scenarios. 
Sec.~\ref{exp:impl_details} outlines datasets, training protocols, and evaluation settings. 
Sec.~\ref{exp:world_generation} assesses world generation, including comparisons with state-of-the-art baselines, analysis under varying camera motions, and ablations on the geometry branch. 
Sec.~\ref{exp:geo_fidelity} examines geometric fidelity through quantitative reconstruction metrics and qualitative visualizations, further validating the role of explicit geometry modeling.

\subsection{Implementation Details}
\label{exp:impl_details}
\textbf{Datasets.}
Our training corpus consists of about 180k video clips collected from a diverse mix of real-world and simulated sources, with geometric supervision obtained through multiple strategies. 
For the RealEstate10K~\citep{zhou2018stereo} and ACID datasets~\citep{infinite_nature_2020}, we generate multi-view consistent depth maps using a reconstruction-based pipeline. 
For additional datasets, which include the real-world DL3DV~\citep{ling2024dl3dv}, WildRGB~\citep{xia2024rgbd}, and ScanNet~\citep{dai2017scannet} together with the simulated TartanAir~\citep{wang2020tartanair}, we apply the Cut3R ~\citep{wang2025continuous} to extract geometric labels.

\textbf{Training Details.}
We train our model with the AdamW optimizer using a learning rate of $10^{-5}$. The process consists of two stages.
In Stage 1 (latent bridging), the geometry branch is trained for 20,000 steps with a global batch size of 64. During this stage, only the geometry branch parameters are updated, adapting it to the feature space of the frozen video backbone. 
In Stage 2 (unified co-optimization), training is performed with 81-frame clips at resolutions of $592 \times 336$ and $336 \times 592$. In this stage, the lightweight interaction modules (bidirectional cross-attention and camera control adapter) are fine-tuned for 10,000 steps with a global batch size of 112, while the core backbones remain frozen. 
We train Stage 1 with 64 H20 GPUs for 36 hours and Stage 2 with 112 H20 GPUs for 144 hours.

\textbf{Evaluation.}
We conduct evaluation on 1,000 samples drawn from the photorealistic subset of the WorldScore static benchmark~\citep{duan2025worldscore}. 
The static split of WorldScore is divided into photorealistic and stylized subsets. 
Since our objective is to model real-world environments for embodied intelligence, we focus on photorealistic video synthesis. 
In contrast, evaluating 3D consistency in stylized videos is ill-defined and less representative of performance in realistic scenarios, and is therefore excluded from our study.
We assess our model along two complementary dimensions: 
\begin{itemize}[leftmargin=*, topsep=2pt, itemsep=0pt, parsep=2pt]
    \item \textbf{World Generation}: For a holistic evaluation of controllable world generation, we report performance on the WorldScore benchmark, which measures performance across multiple axes, including camera and object control, content alignment, 3D consistency, and perceptual quality. 
    \item \textbf{Geometric Fidelity}: To quantify geometric fidelity, we reconstruct each scene with 3DGS, and report PSNR, SSIM, and LPIPS.
\end{itemize}
Together, these metrics capture both the controllability of video-based world generation and the structural accuracy of 3D geometry. For more evaluation details, please refer to \ref{app:eval}.

\subsection{World Generation}
\label{exp:world_generation}

We evaluate the video generation capabilities of \textsc{FantasyWorld} by comparing it against WonderWorld \citep{yu_wonderworld_2025}, Uni3C \cite{cao2025uni3c}, Voyager~\citep{huang_voyager_2025}, and AETHER~\citep{team_aether_2025}, which represent the most relevant state-of-the-art baselines for 3D world generation.

\paragraph{Quantitative Comparison.}
We evaluate under two camera-motion settings. 
The \emph{Small} configuration follows the WorldScore static benchmark, using 1,000 photorealistic samples to ensure consistency with prior work.
We additionally introduce a \emph{Large} setting with 100 curated cases featuring wide orbital or panning trajectories (up to 90°), to test robustness under challenging viewpoints. 
As shown in Table~\ref{tab:worldscore_all}, \textsc{FantasyWorld} achieves the highest scores on all consistency-related metrics (\textit{3D Consist.}, \textit{Photo Consist.}, and \textit{Style Consist.}). 
Moreover, it yields the lowest standard deviation across samples, suggesting that our method is not only more accurate on average but also more stable across diverse scenes.
In the Large setting, the gains are particularly pronounced, demonstrating strong geometric stability even under substantial viewpoint shifts. 
Compared to baselines that prioritize instruction following and camera manipulation \citep{yu_wonderworld_2025}, our framework focuses instead on embedding geometric awareness into video features.
Although this design does not optimize for camera or content alignment benchmarks, it enables stable and reusable 3D representations (our main focus), which the results demonstrate to be both effective and robust.

\begin{figure}[h]
  \centering
  \includegraphics[width=1\linewidth]{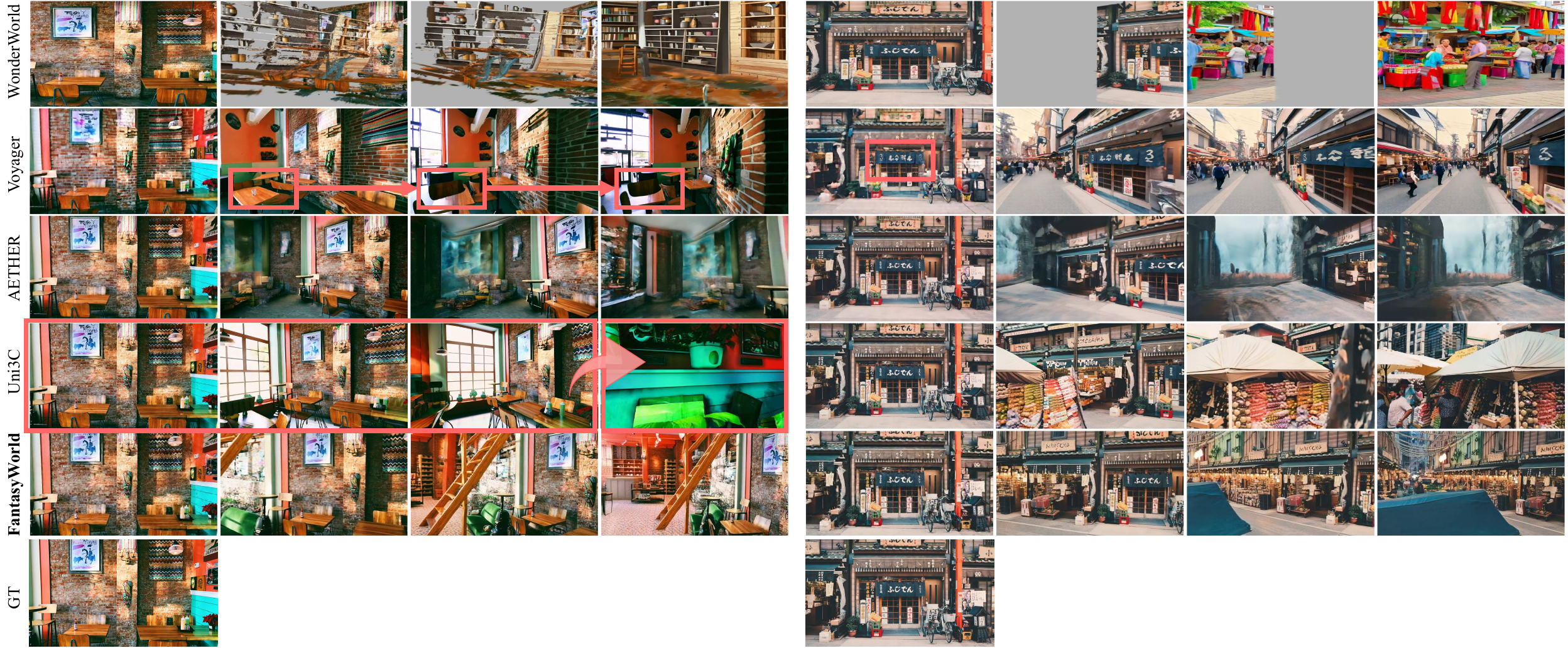}
  \caption{Qualitative comparison of world generation. 
WonderWorld shows missing regions, Voyager suffers from temporal incoherence and degraded first-frame fidelity, AETHER produces low-detail outputs, and Uni3C exhibits abrupt stylistic shifts. 
In contrast, \textsc{FantasyWorld} maintains stronger 3D consistency and coherent style across views.}
  \label{fig:exp_video_gen}
\end{figure}

\textbf{Qualitative Analysis.}
Under large camera motion, distinct failures emerge: WonderWorld shows tearing and holes indicative of missing geometry; in Uni3C and Voyager, first-frame point-cloud priors quickly fall out of view, causing style drift in Uni3C and multi-view misalignment in Voyager; AETHER, despite generating RGB-D point clouds, often produces incoherent, low-detail content (see Fig.~\ref{fig:exp_video_gen}).
In contrast, \textsc{FantasyWorld} predicts an implicit 3D representation that evolves with the video, ensuring stable geometry and appearance across time and yielding coherent, 3D-consistent results without the failures of static priors.

\textbf{Ablation on the Geometry Branch.}
\label{para:ablation}
To assess the role of explicit geometry modeling, we compare our full model with a variant where the geometry branch and bidirectional cross-attention are removed. 
As shown in Table~\ref{tab:worldscore_all}, removing the geometry branch leads to declines in \textit{Photo Consist.} and \textit{Style Consist.}, and an especially severe drop in \textit{3D Consist.}, underscoring its critical role in ensuring multi-view coherence and high-fidelity generation.

\begin{table*}[h!]
\centering
\caption{WorldScore with \emph{Small} vs. \emph{Large} camera motion.}
\label{tab:worldscore_all}
\vspace{2mm}

\small
\setlength{\tabcolsep}{4pt}
\renewcommand{\arraystretch}{1.2}

\begin{tabular}{lc|ccc|cccc}
\toprule
\textbf{Method} & \textbf{Motion} & \textbf{\makecell{3D\\Consist.}} & \textbf{\makecell{Photo\\Consist.}} & \textbf{\makecell{Style\\Consist.}} & \textbf{\makecell{Camera\\Ctrl.}} & \textbf{\makecell{Object\\Ctrl.}} & \textbf{\makecell{Content\\Align.}} & \textbf{\makecell{Subjective\\Qual.}} \\
\midrule
WonderWorld         & Small & 82.85{\tiny$\pm$19.69} & 67.86{\tiny$\pm$23.56} & 55.79{\tiny$\pm$34.89} & \textbf{92.32} & 47.63 & \textbf{79.09} & \textbf{69.03} \\
AETHER              & Small & 79.84{\tiny$\pm$14.68} & 58.68{\tiny$\pm$38.59} & 72.09{\tiny$\pm$32.62} & 57.44 & 52.26 & 28.06 & 41.11 \\
Uni3C               & Small & 78.59{\tiny$\pm$21.08} & 85.48{\tiny$\pm$20.98} & 88.32{\tiny$\pm$18.47} & 62.94 & 45.83 & 47.40 & 57.00 \\
Voyager             & Small & 56.00{\tiny$\pm$26.32} & 80.68{\tiny$\pm$16.32} & 72.89{\tiny$\pm$29.78} & 45.92 & \textbf{57.69} & 48.36 & 44.74 \\
\textbf{Ours w/o 3D} & Small & 79.77{\tiny$\pm$16.06} & 83.86{\tiny$\pm$8.73}  & 92.54{\tiny$\pm$12.90} & 57.94 & 37.33 & 43.31 & 55.85 \\
\textbf{Ours w/ 3D}  & Small & \textbf{83.31{\tiny$\pm$14.24}} & \textbf{86.11{\tiny$\pm$7.97}} & \textbf{94.22{\tiny$\pm$9.11}} & 57.05 & 34.46 & 38.45 & 57.40 \\
\midrule
WonderWorld         & Large & 63.70{\tiny$\pm$24.37} & 3.22{\tiny$\pm$8.47}   & 35.95{\tiny$\pm$33.47} & \textbf{96.28} & 38.61 & \textbf{97.10} & \textbf{72.46} \\
AETHER              & Large & 63.97{\tiny$\pm$17.39} & 33.11{\tiny$\pm$23.99} & 61.99{\tiny$\pm$32.24} & 4.43 & 34.78 & 33.69 & 35.09 \\
Uni3C               & Large & 73.95{\tiny$\pm$17.55} & 46.78{\tiny$\pm$32.64} & 71.43{\tiny$\pm$29.38} & 8.69 & 34.28 & 77.88 & 51.12 \\
Voyager             & Large & 13.82{\tiny$\pm$19.96} & 9.52{\tiny$\pm$17.17}  & 61.34{\tiny$\pm$35.29} & 0.00 & \textbf{49.23} & 64.10 & 39.21 \\
\textbf{Ours w/o 3D} & Large & 72.06{\tiny$\pm$20.14} & 56.98{\tiny$\pm$23.60} & 81.59{\tiny$\pm$22.23} & 9.32 & 34.44 & 75.85 & 46.96 \\
\textbf{Ours w/ 3D}  & Large & \textbf{74.83{\tiny$\pm$16.31}} & \textbf{60.61{\tiny$\pm$21.39}} & \textbf{82.02{\tiny$\pm$19.56}} & 11.24 & 31.96 & 77.20 & 50.46 \\
\bottomrule
\end{tabular}
\end{table*}

\subsection{Geometric Fidelity}
\label{exp:geo_fidelity}
\textbf{Quantitative Comparison.}
We evaluate video–geometry consistency using 3DGS~\citep{kerbl20233d} on 100 \textit{RealEstate10K} samples, comparing three settings (Tab.~\ref{tab:reconstruction_comparison}): removing the geometry branch with VGGT initialization, our full model with VGGT initialization, and our full model with feed-forward point-cloud initialization. 
Results show that, under the same VGGT initialization, adding the geometry branch consistently improves PSNR/SSIM and reduces LPIPS, confirming its role in enforcing 3D consistency.
Direct initialization from our predicted point clouds yields slightly lower scores than VGGT initialization, but still provides competitive results, indicating that our geometry branch produces meaningful 3D structure without relying on external supervision.

\begin{table}[h!]
\centering
\caption{3DGS reconstruction on \textit{RealEstate10K}. Post-reconstruction (Post Rec) indicates the 3DGS initialization source: either from the VGGT point cloud or our own feed-forward point cloud.}


\label{tab:reconstruction_comparison}
\begin{tabular}{l|c|ccc}
\toprule
\textbf{Method} & \textbf{Post Rec.} & \textbf{PSNR} $\uparrow$ & \textbf{SSIM} $\uparrow$ & \textbf{LPIPS} $\downarrow$ \\
\midrule
Ours w/o 3D & VGGT & 26.89 & 0.84 & 0.17 \\
Ours w/ 3D  & VGGT & \textbf{28.24} & \textbf{0.86} & \textbf{0.14} \\
Ours w/ 3D  & - & 26.54 & 0.85 & 0.19 \\
\bottomrule
\end{tabular}
\end{table}

\begin{figure}[h]
\begin{center}
\includegraphics[width=1\linewidth]{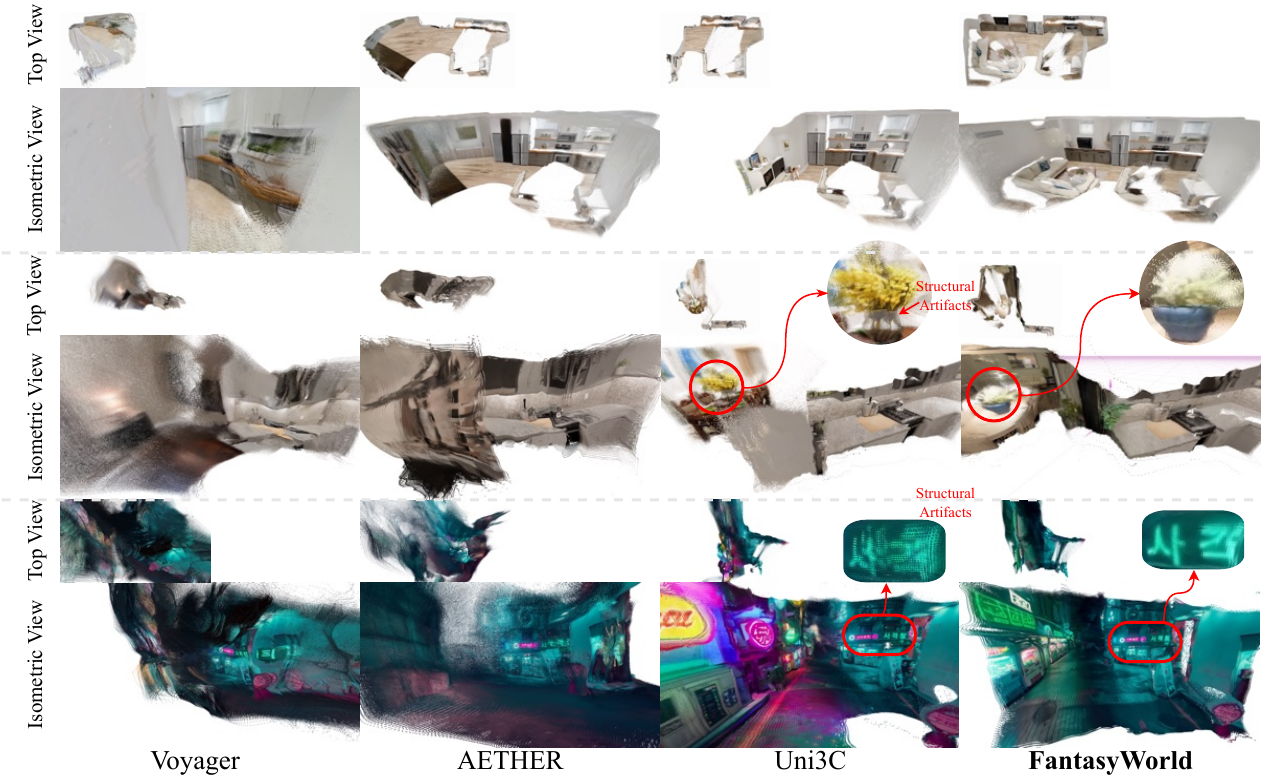}
\end{center}
\caption{Qualitative Comparison of Geometry Fidelity.}
\label{fig:3d_visualize}
\end{figure}

\textbf{Qualitative Analysis.} 
We compare reconstructed 3D scenes from Voyager, AETHER, Uni3C, and our method on indoor and outdoor cases (Fig.~\ref{fig:3d_visualize}). 
All methods use the same pipeline: VGGT predicts point maps from generated frames, then reconstructs point clouds. 
Baselines show structural artifacts and duplication (e.g., blurred signage, layered walls, distorted layouts), while our method yields cleaner, more complete geometry. 
Indoors, walls and corners stay rectilinear and closed; outdoors, text and facades are sharper, indicating stronger 3D consistency.

\textbf{Ablation on the Geometry Branch.}
As shown in Table~\ref{tab:reconstruction_comparison}, the same geometry-ablated variant described in Sec.~\ref{para:ablation} produces weaker reconstruction quality, while the full \textsc{FantasyWorld} achieves clearer geometry and thus confirms the geometry branch is crucial for robust 3D representation.

\section{Conclusion and Limitations}
In this work, we presented \textsc{FantasyWorld}, a unified feed-forward model designed to generate 3D-consistent, explorable virtual worlds in a single pass. 
Our core contribution is a novel architecture that bridges the imaginative power of a pre-trained video diffusion backbone with the geometric rigor of a VGGT-style 3D geometry branch. 
To achieve this, we separate the prediction of 3D geometry from video appearance, introducing a dedicated geometry branch that operates on the video model’s internal features while preserving its powerful generative capabilities.
This is facilitated by a bidirectional cross-attention mechanism that allows video and geometry to mutually reinforce one another during generation. 
Our experiments show that the generated videos not only maintain strong visual realism but also achieve higher multi-view coherence and improved geometric fidelity compared to existing methods, offering a practical and efficient path toward creating structured, reusable world models for embodied AI.

Our model is currently designed for fixed-length clip generation, and extending it to continuous, long-range synthesis remains an important next step. 
Achieving this will require developing effective caching or streaming mechanisms for our implicit 3D representation to maintain state over time. 
Recent work such as Context-as-Memory~\citep{yu2025context} has already begun to address this challenge in the domain of long video generation by introducing memory retrieval strategies that preserve scene consistency across extended trajectories. 
In contrast, our primary focus has been on validating the potential of large video diffusion backbones to serve as world models when paired with geometric supervision. 
This design choice allowed us to clearly demonstrate our central hypothesis: that an implicit 3D representation, learned directly from a video model’s hidden features, is a powerful and effective tool for enforcing 3D consistency.


\section*{Reproducibility Statement}
We ensure reproducibility by detailing \textsc{FantasyWorld}'s architecture (PCB, IRG, bidirectional cross-attention) in Sec.~\ref{sec:method}, dataset composition and preprocessing in Sec.~\ref{exp:impl_details} and~\ref{app:eval}, training protocols in Sec.~\ref{exp:impl_details} and~\ref{app:training}, and evaluation metrics, splits, results, and ablation studies in Sec.~\ref{exp:impl_details},~\ref{exp:world_generation}, and~\ref{exp:geo_fidelity}. 
Baseline code and pretrained weights are documented in~\ref{app:eval}.


\bibliography{iclr2026_conference}
\bibliographystyle{iclr2026_conference}

\appendix
\section{Appendix}

\subsection{Camera Motion Control}
\label{app:camera_control}

For camera motion control, we largely follow the methodology of ~\cite{wan2025wan}, which consists of a camera pose encoder and a pose adapter. Our camera pose encoder mirrors their design, transforming camera parameters into multi-level feature embeddings via Plücker coordinates and a series of convolutional operations.

Our primary modification lies in the camera pose adapter. While the original work utilizes a full Adaptive Layer Normalization (AdaLN) to predict both scaling ($\gamma_i$) and shifting ($\beta_i$) parameters, we propose a streamlined variant that exclusively generates the shifting parameters. The features are integrated into the video latent $f_{i-1}$ at each layer $i$ through a simple additive projection:
\[
f_i = f_{i-1} + \beta_i,
\]
where $\beta_i$ is derived from the camera embeddings. Also the camera control module is adapt only the first 24 block among the 40 transformer blocks.
\subsection{3D DPT Head Architecture}
\label{app:dpt3d}

\begin{figure}[h]
  \centering
  \includegraphics[width=1\linewidth]{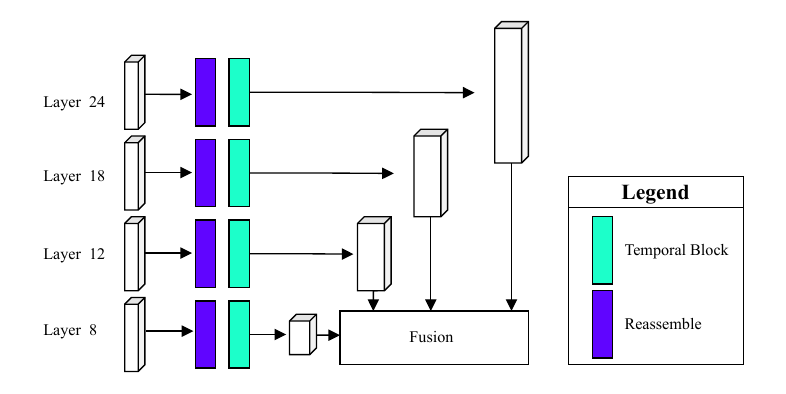}
  \caption{Our 3D DPT head with temporal blocks for temporal upsampling.}
  \label{fig:appendix_3d_dpt}
\end{figure}

Our model employs a custom 3D DPT head, which extends the 2D version from VGGT~\citep{wang2025vggt} to process video sequences via two key innovations, as shown in Fig \ref{fig:appendix_3d_dpt}.

First, we invert the spatial reassembly logic to align with the nature of diffusion backbones. 
Conventional DPTs, designed for standard encoders, assume that shallow layers contain high-frequency spatial details and therefore upsample them most aggressively. 
Diffusion models operate differently: they progressively denoise the input at each block. 
This means deeper blocks produce features that are not only semantically richer but also less noisy and more structurally reliable. 
Consequently, we invert the DPT logic. 
Our Geometry-Consistent Branch uses features from blocks \{8,12,18,24\} and upsamples features from the deepest layers the most, while downsampling those from the shallower ones. 
This anchors the fusion process in the highest-quality information the backbone provides.

Second, we temporally upsample each feature stream using two sequential Temporal Blocks after the reassemble block. The design of the upsample block is inspired by the WanVAE decoder~\citep{wan2025wan}. Each block first doubles the temporal resolution and then applies a causal 3D convolution. This process results in a total 4x temporal upsampling factor, transforming an input of $t$ frames into a smooth output sequence of $T = 4(t-1)+1$ frames.

\subsection{Training Details for the Geometry Branch}
\label{app:training}

\textbf{Camera Pose Prediction.}
Following VGGT~\citep{wang2025vggt}, we concatenate one learned camera token and four register tokens to the video token sequence. 
Because camera trajectories are temporally smooth in videos, we apply a lightweight 1D convolution to the camera token stream to perform temporal upsampling. 
For each frame $f_i$, the geometry head predicts a 9D camera parameter vector
$\mathbf{g}_i \in \mathbb{R}^{9}$ (rotation, translation, and field-of-view).

\textbf{Depth and Point Map prediction}
We use hidden features from the Geometry-Consistent Branch's layers $\{8,12,18,24\}$ with shapes
$\mathbb{R}^{t \times (h\!\times\!w) \times c}$.
The features are fed a Depth 3D DPT head and a Point-Map 3D DPT head.
The depth head produces $D \in \mathbb{R}^{T \times H \times W}$, and the point head produces a 3D point map
$P \in \mathbb{R}^{T \times H \times W \times 3}$ together with a confidence (uncertainty) map
$\Sigma^{P} \in \mathbb{R}^{T \times H \times W}$, where $h=\frac{H}{16}$, $w=\frac{W}{16}$, and $T=(t-1)\times 4 + 1$.

\textbf{Camera loss.}
We supervise camera parameters with a robust Huber loss, where $\hat{\mathbf{g}}_i $ is the ground truth:
\[
\mathcal{L}_{\text{camera}}
\,=\, \sum_{i=1}^{N} \bigl\lVert \hat{\mathbf{g}}_i - \mathbf{g}_i \bigr\rVert_{\epsilon}.
\]

\textbf{Depth loss.}
We adapt Video Depth Anything~\citep{chen2025video} and combine a temporal gradient matching term with a per-frame scale-sensitive spatial loss term:
\[
\mathcal{L}_{\text{depth}}
\,=\, \alpha\, \mathcal{L}_{\text{TGM}} \;+\; \beta\, \mathcal{L}_{\text{frame}},
\]
where $\mathcal{L}_{\text{TGM}}$ enforces temporal consistency of depth gradients, and $\mathcal{L}_{\text{frame}}$ measures per-frame depth error without scale/shift normalization.

\textbf{Point-map loss.}
Following VGGT (\cite{wang2025vggt}), we penalize both point positions and local gradients, weighted by the predicted uncertainty:
\[
\mathcal{L}_{\text{pmap}}
\,=\, \sum_{i=1}^{N}
\Bigl\lVert \Sigma_i^{P} \odot \bigl(\hat{P}_i - P_i\bigr) \Bigr\rVert
\;+\;
\Bigl\lVert \Sigma_i^{P} \odot \bigl(\nabla \hat{P}_i - \nabla P_i\bigr) \Bigr\rVert
\;-\; \gamma \,\log \Sigma_i^{P}.
\]

\textbf{Total objective.}
The geometry branch is trained with
\[
\mathcal{L}_{\mathrm{geo}}
\,=\, \mathcal{L}_{\mathrm{depth}}
\;+\; \mathcal{L}_{\mathrm{pmap}}
\;+\; 3\,\mathcal{L}_{\mathrm{camera}}.
\]

\subsection{Evaluation Details}
\label{app:eval}

To ensure the reproducibility of our evaluation, we provide the implementation details of all baselines considered in our benchmark. For each method, we list the official GitHub repository, the specific commit used in our experiments, and the resolution and number of frames generated per sample.

\textbf{Repositories and Commits.}
\begin{itemize}[leftmargin=*, topsep=2pt, itemsep=2pt, parsep=2pt]
  \item \textbf{Voyager:} \url{https://github.com/Tencent-Hunyuan/HunyuanWorld-Voyager} \\(commit \texttt{54a658b})
  \item \textbf{Uni3C:} \url{https://github.com/alibaba-damo-academy/Uni3C} (commit \texttt{75ed6e2})
  \item \textbf{AETHER:} \url{https://github.com/InternRobotics/Aether} (commit \texttt{f2221b8})
  \item \textbf{WonderWorld:} \url{https://github.com/KovenYu/WonderWorld} (commit \texttt{5cf1146})
\end{itemize}

\textbf{Resolution and Frame Settings.}  
We followed the official default inference settings of each baseline:  
\begin{itemize}[leftmargin=*, topsep=2pt, itemsep=2pt, parsep=2pt]
    \item \textbf{FantasyWorld (ours):} $336 \times 592$, $81$ frames  
    \item \textbf{Voyager:} $512 \times 768$, $49$ frames  
    \item \textbf{Uni3C:} $480 \times 768$, $81$ frames  
    \item \textbf{AETHER:} $480 \times 720$, $41$ frames  
    \item \textbf{WonderWorld:} $512 \times 512$, $50$ frames
\end{itemize}

All baselines were run with the official inference scripts and default hyperparameters from the respective repositories.


\end{document}